\documentclass[nohyperref]{article}
\usepackage{PRIMEarxiv}

\usepackage[utf8]{inputenc} 
\usepackage[T1]{fontenc}    
\usepackage{hyperref}       
\usepackage{url}            
\usepackage{booktabs}       
\usepackage{amsfonts}       
\usepackage{nicefrac}       
\usepackage{microtype}      
\usepackage{lipsum}
\usepackage{fancyhdr}       
\usepackage{graphicx}       
\graphicspath{{media/}}     

\title{Cyclical Focal Loss
}

\author{
	Leslie N. Smith \\
	U.S. Naval Research Laboratory, \\
	Naval Center for Applied Research in AI, Code 5514 \\
	4555 Overlook Ave., SW., Washington, D.C.  20375 \\
	\texttt{leslie.smith@nrl.navy.mil} \\
}

\usepackage{hyperref}

\usepackage{amsmath}
\usepackage{amssymb}
\usepackage{mathtools}
\usepackage{amsthm}
\usepackage{listings}
\usepackage{color}

\newcommand{\etal}{et~al. }

\usepackage[capitalize,noabbrev]{cleveref}

\theoremstyle{plain}

\theoremstyle{definition}

\theoremstyle{remark}

\usepackage[textsize=tiny]{todonotes}

\begin{document}
\maketitle

\begin{abstract}
The cross-entropy softmax loss is the primary loss function used to train deep neural networks.  On the other hand, the focal loss function has been demonstrated to provide improved performance when there is an imbalance in the number of training samples in each class, such as in long-tailed datasets.  In this paper, we introduce a novel cyclical focal loss and demonstrate that it is a more universal loss function than cross-entropy softmax loss or focal loss.  We describe the intuition behind the cyclical focal loss and our experiments provide evidence that cyclical focal loss provides superior performance for balanced, imbalanced, or long-tailed datasets.  We provide numerous experimental results for CIFAR-10/CIFAR-100, ImageNet, balanced and imbalanced 4,000 training sample versions of CIFAR-10/CIFAR-100, and ImageNet-LT and Places-LT from the Open Long-Tailed Recognition (OLTR) challenge.  Implementing the cyclical focal loss function requires only a few lines of code and does not increase training time.  
In the spirit of reproducibility, our code is available at \url{https://github.com/lnsmith54/CFL}.
\end{abstract}

\section{Introduction}
\label{sec:intro}

The use of trained neural networks is pervasive in a wide variety of societal applications such as medical diagnosis, scientific discovery, and the defense of our homes and country.  In the majority of cases, a cross-entropy softmax loss guides the training of networks. 

\begin{figure}[tb]
	\begin{center}
		\centerline{\includegraphics[width=0.8\columnwidth]{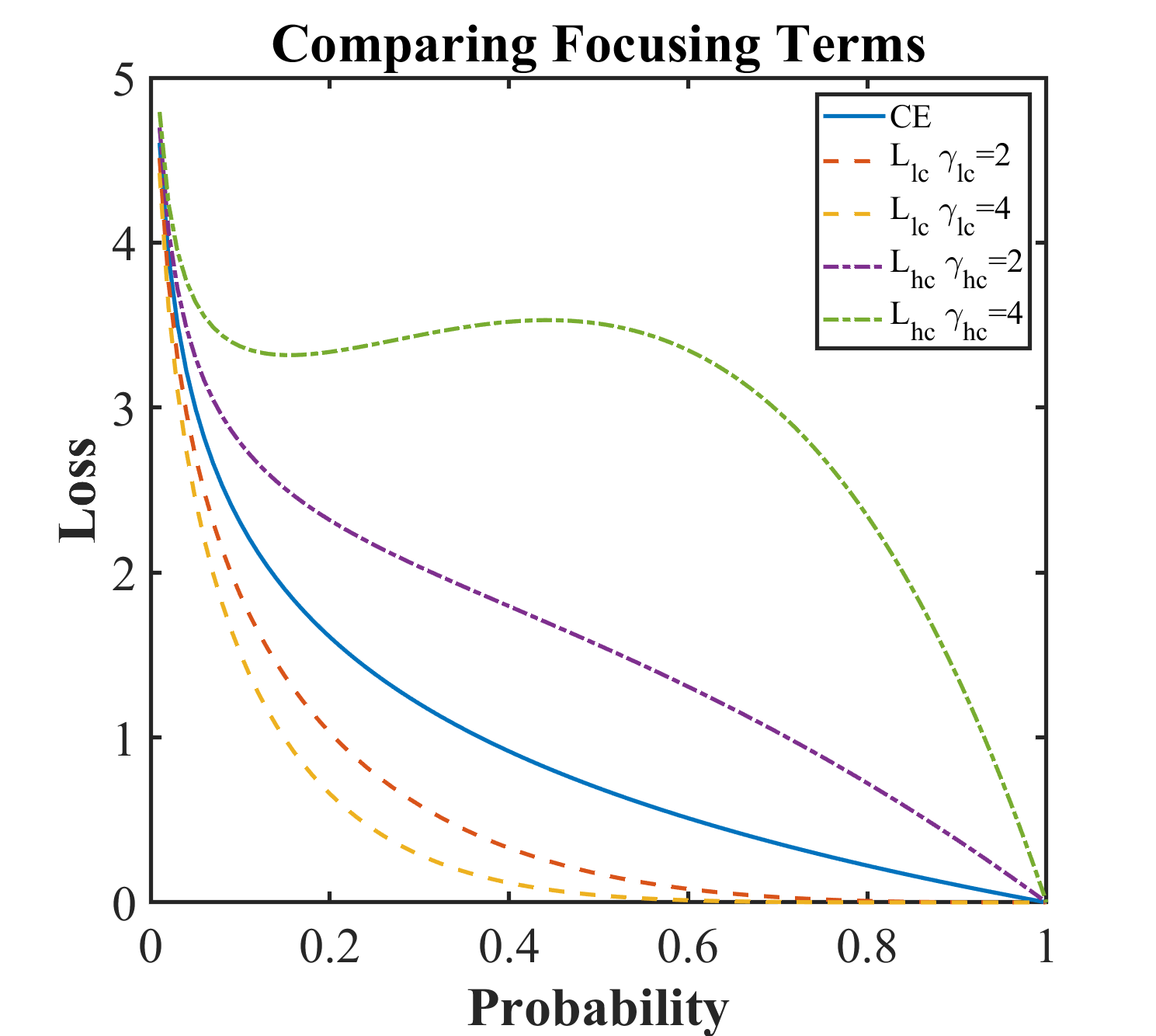}}
		\caption{\textbf{Loss Weighting Factors:} This figure compares the loss weighting factors versus cross-entropy softmax (CE) for focal loss (FL) and our new high confidence weighting term ($L_{hc}$) as a function of the ground truth probability $p_t$.  The CE loss is $  -log( p_t )$, the FL loss term is $ -(1 - p)^{\gamma_{lc}} log( p_t )$, and the $L_{hc}$ term is $ -(1 + p)^{\gamma_{hc}} log( p_t )$.
		}
		\label{fig:gamma}
	\end{center}
	\vskip -0.2in
\end{figure}

On the other hand, focal loss \cite{lin2017focal} is superior to the cross-entropy softmax loss when there is an imbalance, such as in the number of training samples per class, a foreground-background imbalance \cite{lin2017focal}, or a positive-negative label imbalance in multi-label classification \cite{ridnik2021asymmetric}.
Focal loss  modifies the cross-entropy softmax loss to increase the focus of a neural network's training on the hard, misclassified data samples.  That is, the focal loss is governed by:
\begin{equation}
	L_{lc} = - (1 - p_t)^{\gamma_{lc}} log( p_t ),
	\label{eqn:FLterm}
\end{equation}
where $L_{lc}$ focuses the loss on the low confidence samples, $p_t$ is the softmax probabilities, and focal loss adds the weight $(1 - p_t)^{\gamma_{lc}}$ to the cross-entropy softmax loss.  As the probability $p_t$ goes to 1 for more confident training samples, the weight for the loss of that training sample drives the loss term to zero faster than for cross-entropy, as shown in Figure \ref{fig:gamma}.  The impact of this weighting is to focus the network training on the rarer and less confident training samples.  
When $\gamma_{lc} = 0$, the focal loss becomes identical to the cross-entropy softmax loss.

While the focal loss has been found beneficial in tasks with imbalanced class data, it generally reduces the performance when training with more balanced datasets.  Therefore, the focal loss is not a good replacement for the cross-entropy softmax loss for most applications.

A paper on the concept of “General Cyclical Training of Neural Networks” \cite{smith2022general} defines the general cyclical training of neural networks as network training that starts and ends with a focus on the easy and confident training samples and trains during the middle epochs with an increasing and then decreasing focus on the hard training samples. 
In other words, general cyclical training can be considered as a combination of curriculum learning \cite{bengio2009curriculum} in the early epochs, training on the full problem space happening during the middle epochs, and fine-tuning on confident samples at the end.  

In this paper, we propose a cyclical focal loss (CFL) that follows this principle of general cyclical training.  Specifically, we propose a new loss weighting term as
\begin{equation}
	 L_{hc} = - (1 + p_t)^{\gamma_{hc}} log( p_t ),
	\label{eqn:CFLterm}
\end{equation}
where this term causes the loss to increase the focus on the more confident training samples (named $L_{hc}$ for focusing on the high confidence training samples).  Figure \ref{fig:gamma} compares the weighting terms for $\gamma_{hc} = 2$ and 4 in Equation \ref{eqn:CFLterm} to the standard cross-entropy.
It is clear in this Figure that this focusing term weights the loss of samples for which $p_t$ is close to 1 substantially more than cross-entropy does.
We define the cyclical focal loss function so that more confident training samples are weighted more heavily in the early and final epochs via this new term, while in the middle epochs, the less confident training samples are more heavily weighted via Equation \ref{eqn:FLterm}.   

An additional inspiration for cyclical focal loss comes from curriculum learning \cite{bengio2009curriculum}. 
Curriculum learning implies that it is best to construct a neural network training methodology for the task at hand such that the network’s weight updates during earliest epochs are encouraged by the confident training samples via Equation \ref{eqn:CFLterm}.  The training of hard samples is performed in the middle epochs to improve generalization.  At the end of the training is a fine-tuning stage on the confident samples learned by the network, because this is when the network learns the more complex patterns \cite{you2019does} from the most confident training samples. 

This paper demonstrates that cyclical focal loss is a more universal loss function than cross-entropy softmax loss or focal loss for balanced or imbalanced datasets.   In our experiments on balanced datasets, we show that using CFL generally improves on the network’s generalization performance, and at worst is comparable to the cross-entropy softmax loss.  Our experiments with only 4,000 training samples from CIFAR-10 and CIFAR-100 show superior performance for CFL for both balanced and imbalanced versions.
We also demonstrate in the open long-tailed recognition (OLTR) challenge \cite{liu2019large} that the cyclical focal loss improves on the network's performance more than training with either softmax or focal loss.  These experiments provide evidence of CFL's superiority across balanced, imbalanced, or highly imbalanced datasets.

Our contributions are:
\begin{itemize}
	\item We propose a novel loss function, the cyclical focal loss, that begins and ends training with a focus on the confident training samples and trains in the middle epochs with a focus on the hard training samples. 
	\item We demonstrate that the cyclical focal loss improves on the performance of the trained network over training with the standard cross-entropy softmax and the focal loss for CIFAR-10/100 (with balanced, imbalanced, or limited data training data), ImageNet, and in the open long tail recognition problem.
	\item Therefore, our experiments demonstrate that cyclical focal loss is a more universal loss function that often performs better than and can replace the standard cross-entropy softmax loss or focal loss.
	\item Our implementation does not increase training or inference time. We describe our implementation in the Appendix and share our fully reproducible training code is available at \url{https://github.com/lnsmith54/CFL}.
\end{itemize}


\section{Related Work}
\label{sec:related}

\textbf{Focal loss function:}
The focal loss function was first introduced for object detection \cite{lin2017focal}.  These authors discovered that extreme foreground-background imbalance was the cause of the inferior performance of 1-stage detectors and showed that their proposed focal loss function improved the performance of these detectors.  The focal loss heavily weights less confident training samples, as shown in Figure \ref{fig:gamma}.  After the introduction of focal loss, others have leveraged the focal loss in other situations of imbalance \cite{ridnik2021asymmetric,li2020generalized,spiegl2021contrastive,mukhoti2020calibrating,yun2019focal}.

Focal loss was applied to multi-label classification by Ridnik, et al. \cite{ridnik2021asymmetric}, where there is a positive label/negative label imbalance; that is, in a given image, most of the labels are not present, so the negative label examples are much more prevalent than the positive labels.  These author's proposed a novel variation of focal loss, which they called ``asymmetric loss'' (ASL), that gave improved performance over the original focal loss for multi-label classification.  In ASL, the positive and negative terms of the focal loss are separated and each have its own focusing parameter.

\textbf{General cyclical training:}
General cyclical training was defined \cite{smith2022general} as any collection of settings in machine learning where the training starts and ends with ``easy training'' and the ``hard training'' happens during the middle epochs.
General cyclical training  can be considered as a combination of curriculum learning \cite{bengio2009curriculum} in the early epochs with fine-tuning toward the end of training, plus training on the full problem space during the middle epochs. 
It has been shown that many important aspects of neural network learning take place within the very earliest iterations or epochs of training  \cite{golatkar2019time,frankle2020early}.  It is best to start a neural network's training with highly confident samples to encourage the network’s weight updates in an optimal direction.  As the training proceeds, increasing the variation and range of the training and data improves the generalization of the model.  While this first part of a network’s training can use a curriculum learning approach, the last epochs of the training should fine-tune the model for the desired data and task in order to to encourage the network to learn the more complex patterns \cite{you2019does} from the most confident training samples.   

Adaptive hyper-parameters during training have become common.  Cyclical learning rates \cite{smith2017cyclical,smith2019super,loshchilov2016sgdr} have been accepted by the deep learning community.  In addition, the commonly used learning rate warmup \cite{goyal2017accurate} and stochastic gradient descent (SGDR) with restarts \cite{loshchilov2016sgdr} are essentially equivalent to cyclical learning rates.  Furthermore, the idea of adaptive hyper-parameters has been extended to other hyper-parameters, such as weight decay \cite{zhang2018three,bjorck2020understanding,nakamura2019adaptive,lewkowycz2020training} and batch sizes \cite{smith2017don}.  
In this paper, we extend the cyclical training principle to loss functions by proposing the cyclical focal loss.

\section{Methods}
\label{sec:methods}

\subsection{Focal Loss}
\label{subsec:FL}

Following Lin, \etal \cite{lin2017focal}, we start with the cross-entropy loss for binary classification as:
\begin{equation}
	CE(p, y) =
	\begin{cases}
		- log( p )  & \quad \text{if } y = 1 \\
		- log( 1 - p )  & \quad \text{otherwise} 
	\end{cases}
	\label{eqn:CE}
\end{equation}
where $ y $ specifies ground truth class and $ p \in [0, 1] $ is the model's estimated probability.  For simplicity, these authors define:
\begin{equation}
	p_t =
	\begin{cases}
		 p  & \quad \text{if } y = 1 \\
		 1 - p   & \quad \text{otherwise}, 
	\end{cases}
	\label{eqn:p_t}
\end{equation}
so that CE can be written at $ CE(p, y) = CE( p_t ) = - log( p_t ) $.

Using this notation, the focal loss function was defined as:
\begin{equation}
	FL( p_t ) = -( 1 - p_t )^\gamma log( p_t ),
	\label{eqn:FL}
\end{equation}
where $\gamma \geq 0$ is a tunable hyper-parameter set by the user. As seen in Figure \ref{fig:gamma}, the weighting factor reduces the loss contribution for confident predictions, which increases the importance of correcting misclassified samples.  Note that when $\gamma = 0$, the focal loss is equivalent to the cross-entropy loss.  Equation \ref{eqn:FL} is the same as Equation \ref{eqn:FLterm} but we rename $\gamma$ as $\gamma_{lc}$ for clarity in the rest of this paper.

Recently, Ridnik, \etal \cite{ridnik2021asymmetric} generalized the focal loss to improve on multi-label classification, in which the number of negative labels in a training sample greatly exceeds the number of positive labels (i.e., imbalanced data).  These authors propose decoupling the weighting factor between the positive and negative labels by having separate $\gamma$ hyper-parameters for the positive and negative parts of the loss.  Therefore, they define an asymmetric loss (ASL) as:
\begin{equation}
	ASL(p, y) = L_+ + L_-,
	\label{eqn:ASL}
\end{equation}
where
\begin{equation}
	\begin{cases}
		L_+ = - (1 - p) ^{\gamma_+} log( p ) \\
		L_- = - (p)^{\gamma_-} log( 1 - p ).
	\end{cases}
	\label{eqn:Lpm}
\end{equation}
The authors replace the one user-defined hyper-parameter of $\gamma$ with two hyper-parameters $\gamma_+$ and $\gamma_-$.
They note that $\gamma_- > \gamma_+$, and based on their experiments, suggest setting $\gamma_+ = 0$ so that the positive examples use the cross entropy loss.  In this case, ASL differs from cross entropy only in the weighting of the negative labels.  

Next, we define cyclical versions of both the focal loss and the asymmetric focal loss.

\subsection{Cyclical Focal Loss}
\label{subsec:CFL}

Intuitively, the goal for the cyclical focal loss is to combine a focus on confident predictions in the early epochs with an increasing focus on misclassified, hard samples during the middle epochs.  We can accomplish this by including both the loss terms in Equations \ref{eqn:FLterm} and \ref{eqn:CFLterm} with any reasonable schedule between them.  For simplicity, we use a linear schedule in this paper.

That is, we define a parameter $\xi$ that varies with the training epoch as:
\begin{equation}
	\xi = \begin{cases}
		1 - f_c \frac{e_i}{e_n}      & \quad \text{if } f_c \times e_i \leq e_n \\
		\big( f_c \frac{e_i}{e_n} - 1\big)  / (f_c - 1)  & \quad \text{ otherwise}
	\end{cases}
	\label{eqn:xi}
\end{equation}
where $e_i$ corresponds to the current training epoch number and $e_n$ corresponds to the total number of training epochs.
Here, we introduce a cyclical factor $f_c \geq 1$ that provides variability to the cyclical schedule. 
If $f_c = 1, \xi$ goes from 1 at the beginning of the training to 0 at the end.   If the cyclical factor $f_c = 2$, the cycle resembles an upside down equilateral triangle, where $\xi$ goes from 1 to 0 in the first half of the epochs and from 0 to 1 in the second half.
If $f_c = 4, \xi$ goes from 1 to 0 at a quarter of the way through the training and linearly goes from 0 to 1 for the remaining three-quarters of the epochs.

Integrating Equations \ref{eqn:FLterm} and \ref{eqn:CFLterm} with Equation \ref{eqn:xi}, we define the cyclical focal loss as:
\begin{equation}
	CFL(p, y ) = \xi L_{hc} + (1 - \xi) L_{lc}
	\label{eqn:CFL1}
\end{equation}
Our cyclical focal loss introduces two new hyper-parameters, $f_c$ and $\gamma_{hc}$ but we show in our experiments that this is not a hardship.  Specifically, $\gamma_{hc} = 2$ or 3 generally works well and the results are fairly insensitive to the value of $f_c$.  We used $f_c = 4$ throughout our experiments.

Similarly, we also tested a cyclical version of the asymmetric loss (which is defined in Equation \ref{eqn:ASL}).  
Our cyclical asymmetric focal loss is defined as:
\begin{equation}
	CASL(p, y ) = \xi L_{hc} +  (1 - \xi) ( L_+ + L_-)
	\label{eqn:CFL2}
\end{equation}
where $L_+$ and $L_-$ are defined in Equation \ref{eqn:Lpm}.  
While our cyclical asymmetric focal loss is not symmetric between the high-confidence and low confidence terms, this does incorporate our goal to train on the confident samples early and late in the training.
In our experiments with this cyclical asymmetric focal loss, we used $\gamma_- = 4 $ and $ \gamma_+ =0$, as recommended in the original paper for ASL.

We mention that our implementation of cyclical loss functions is simple and requires only a few lines of code.  Our implementation does not increase training time. More details on the implementation are provided in the Appendix and we share our training code is available at \url{https://github.com/lnsmith54/CFL}.


\begin{table*}[tb]
	\caption{\textbf{Comparisons with balanced datasets:} Comparison of the top-1 test classification accuracies for cross-entropy softmax (CE), focal loss (FL), asymmetric loss (ASL), cyclical focal loss (CFL), and cyclical asymmetric focal loss (CASL). Both cyclical focal loss results are comparable or better than the results from training with cross-entropy softmax, the focal loss, and the asymmetric focal loss across all of these experiments. Reported accuracies are the mean and the standard deviation of four runs for CIFAR and of two runs for ImageNet. The best results are highlighted in bold. }
	\label{tab:losses}
	\begin{center}
		\begin{small}
			\begin{sc}
				\begin{tabular}{|l|c|c|c|c|c|}
					\hline
					Data set / Model & CE & FL & ASL  & CFL (Ours)  & CASL (Ours) \\
					\hline
					CIFAR-10/TResNet\_m  & 97.32$\pm$ 0.06& 96.88$\pm$ 0.08&  97.3$\pm$ 0.08& \textbf{97.40$\pm$ 0.08} &  97.35$\pm$ 0.05 \\
					\hline
					CIFAR-10/ResNet50    & 96.41$\pm$ 0.09& 95.65$\pm$ 0.05&  96.25$\pm$ 0.01& \textbf{96.88$\pm$ 0.01} &  96.69.$\pm$ 0.04 \\
					\hline
					CIFAR-10/Efficient\_B0    & \textbf{97.50$\pm$ 0.11}  & 97.16$\pm$ 0.11&  97.48$\pm$ 0.06& 97.47$\pm$ 0.08&  \textbf{97.51$\pm$ 0.06} \\
					\hline
					CIFAR-100/TResNet\_m  & 83.95$\pm$ 0.20& 83.33$\pm$ 0.07&  83.89$\pm$ 0.19& 84.22$\pm$ 0.06 &  \textbf{84.24$\pm$ 0.20} \\
					\hline
					ImageNet/TResNet\_m  & 80.03$\pm$ 0.08& 78.34$\pm$ 0.07&  79.20$\pm$ 0.01& \textbf{80.45$\pm$ 0.02} &  80.33$\pm$ 0.12 \\
					\hline
				\end{tabular}
			\end{sc}
		\end{small}
	\end{center}
	\vskip -0.1in
\end{table*}

\section{Experiments}
\label{sec:esp}

We now validate the performance of both the cyclical focal loss (CFL) and the cyclical asymmetric loss (CASL).
In this Section we provide results from  experiments to compare our loss functions to cross-entropy softmax (CE), focal loss (FL), and asymmetric focal loss (ASL).  We also provide experimental results to better understand the new hyper-parameters: $\gamma_{hc}$ and the cyclical factor $f_c$.  We demonstrate that the performance with the cyclical loss functions are comparable or better than cross-entropy and focal loss for a variety of datasets and whether the training datasets are balanced or imbalanced.

\textbf{Datasets:}
We present results for the following datasets: CIFAR-10/CIFAR-100 \cite{krizhevsky2009learning}, ImageNet \cite{deng2009imagenet}, balanced and imbalanced 4,000 training sample versions of CIFAR-10 and CIFAR-100, and ImageNet-LT and Places-LT for the Open Long-Tailed Recognition dataset (OLTR) \cite{liu2019large}.
These benchmark datasets provide a range of classification problems, including a highly imbalanced scenario with ImageNet-LT/Places-LT.  

\textbf{Model and hyper-parameters:}
Details of the implementations and hyper-parameters are given in the Appendix.
In a majority of our experiments we used a TResNet \cite{ridnik2021tresnet} but we also show results with the ResNet-50 and EfficientNet\_B0 \cite{tan2019efficientnet} architectures

We modified the implementation of the asymmetric focal loss\footnote{\url{https://github.com/Alibaba-MIIL/ASL}} \cite{ridnik2021asymmetric} to include a cyclical focal loss option.  The asymmetric focal loss provided the flexibility to compare cyclical versions to both the original and asymmetric focal losses.

For the open long tailed recognition experiments \cite{liu2019large}, we modified the implementation provided by the authors\footnote{\url{https://github.com/zhmiao/OpenLongTailRecognition-OLTR}} to use focal loss and our cyclical focal loss.

For all our other experiments and datasets, we used PyTorch Image Models (TIMM)\footnote{\url{https://github.com/rwightman/pytorch-image-models}} \cite{rw2019timm} as a framework in our experiments.  This framework provides the models  used in our our experiments.
To ease replication of our work, code with modifications to this framework is provided at \url{https://github.com/lnsmith54/CFL}.

\begin{figure}[htb]
	\vskip 0.2in
	\begin{center}
		\centerline{\includegraphics[width=0.8\columnwidth]{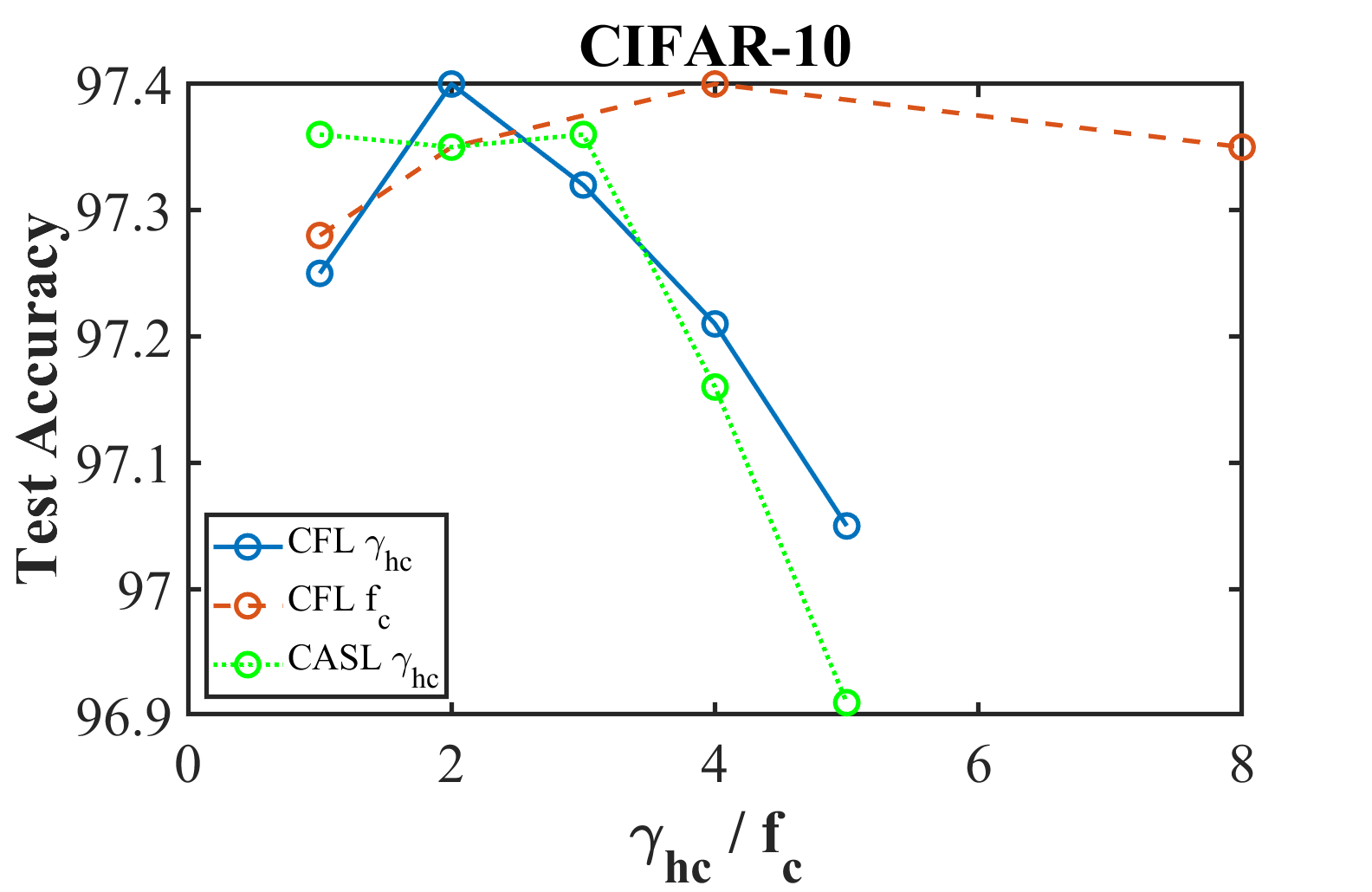}}
		\caption{\textbf{$\gamma_{hc} \ f_c$:} This figure shows the impact on CIFAR-10 test accuracy for a range of values for $\gamma_{hc}$ and $f_c$. For cyclical focal loss, the best value for $\gamma_{hc}$ is 2 and the accuracy drops for higher values of $\gamma_{hc}$.  For cyclical asymmetric focal loss, training within the range of $\gamma_{hc}$ from 1 to 3 gives accuracies that are nearly constant and within the precision of our experiments.
			For the $f_c$ parameter the accuracy changes only slightly over a range of settings, and mainly falls within the precision of our experiments (by default, we used $f_c =  4$ in our experiments).
		}
		\label{fig:c10hp}
	\end{center}
	\vskip -0.2in
\end{figure}

\subsection{CIFAR-10 and CIFAR-100}
\label{subsec:cifar}

This Section evaluates CFL's and CASL's test accuracy results for both CIFAR-10 and CIFAR-100.

Table \ref{tab:losses} shows the test accuracies comparing cross-entropy softmax (CE), focal loss (FL) \cite{lin2017focal}, asymmetric focal loss (ASL) \cite{ridnik2021asymmetric}, cyclical focal loss (CFL), and cyclical asymmetric focal loss (CASL).
For CIFAR-10 and CIFAR-100, each entry of test accuracy in this Table is the mean and the standard deviation of four runs with the same hyper-parameters and loss function. 

The first row of Table \ref{tab:losses} gives the results for CIFAR-10 with the TResNet\_m architecture. In the second row are the results for a ResNet-50 model.  The third row presents the results with the EfficientNet\_B0 architecture.  The fourth row of this Table gives the results for CIFAR-100 with the TResNet\_m architecture.
It is clear from the third column of this Table that using the original focal loss on these datasets reduces the performance of the network for CIFAR-10 and CIFAR-100 in all four cases.  However, using the asymmetric focal loss (see column 4) generally regains much of the lost performance.  This is to be expected because $\gamma_+ = 0$  is the same as the cross entropy loss for the positive labels.  We note that the implication is that the loss contribution from the error or negative labels is minor for CIFAR-10 and CIFAR-100 classification.

On the other hand, the fifth and sixth columns of Table \ref{tab:losses} show that the test accuracy when training with our cyclical focal loss and the cyclical asymmetric focal loss is consistently better than when training with the other loss functions. Note that the results for cross-entropy softmax with the EfficientNet\_B0 are higher than the performance results for TResNet\_m and ResNet-50 and that using our cyclical focal loss functions only gives comparable results.  This implies that in situations where near optimal performance is being reached (i.e., optimal hyper-parameters), the cyclical focal loss might not improve on the network's performance, but it also does not harm it.  On the other hand, throughout all of our experiments, our cyclical focal loss functions provided near-optimal performance with less need of precision search for optimal hyper-parameters.  That is, we found that training with cyclical focal loss reduced the need for tedious hyper-parameter searches.

Figure \ref{fig:c10hp} shows the test accuracies for CIFAR-10 with TResNet\_m over a range of values for $\gamma_{hc}$ and $ f_c$ for CFL and CASL.  For CFL, the best value for $\gamma_{hc}$ is 2, but using $\gamma_{hc} = 3$ is within the precision of these tests (i.e., approximately $\pm0.1$).  For CASL, the best value for $\gamma_{hc}$ is 1 but values of 2 or 3 are within the precision of our experiments.  
The test accuracies were little changed over a range of values of $ f_c$ for CFL and CASL (not shown) and we used a value of $f_c = 4$ in our experiments.  A similar plot was obtained for CIFAR-100 and is not shown.


\begin{figure}[htb]
	\begin{center}
		\centerline{\includegraphics[width=0.75\columnwidth]{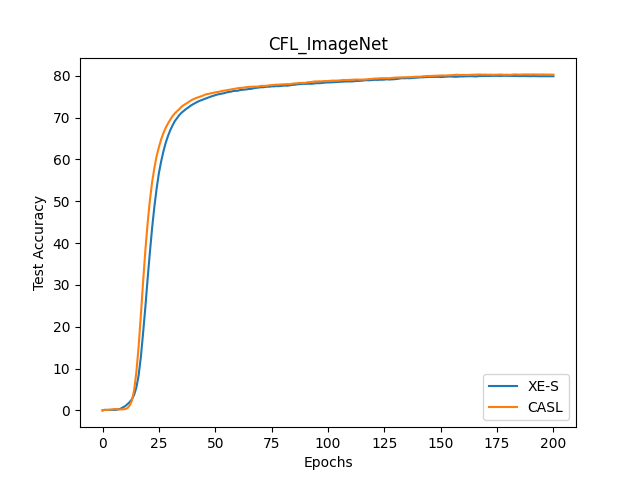}}
		\caption{\textbf{ImageNet training curves:} This figure shows the ImageNet test accuracy over the course of training with a TResnet\_m model for cross-entropy softmax (CE) and our cyclical asymmetric focal loss (CASL).  The two curves are very similar but the faster learning curve for CASL supports our assertion that cyclical focal loss makes learning easier in the early epochs.
		}
		\label{fig:imagenet_training}
	\end{center}
	\vskip -0.2in
\end{figure}
\begin{figure}[htb]
	\begin{center}
		\centerline{\includegraphics[width=0.75\columnwidth]{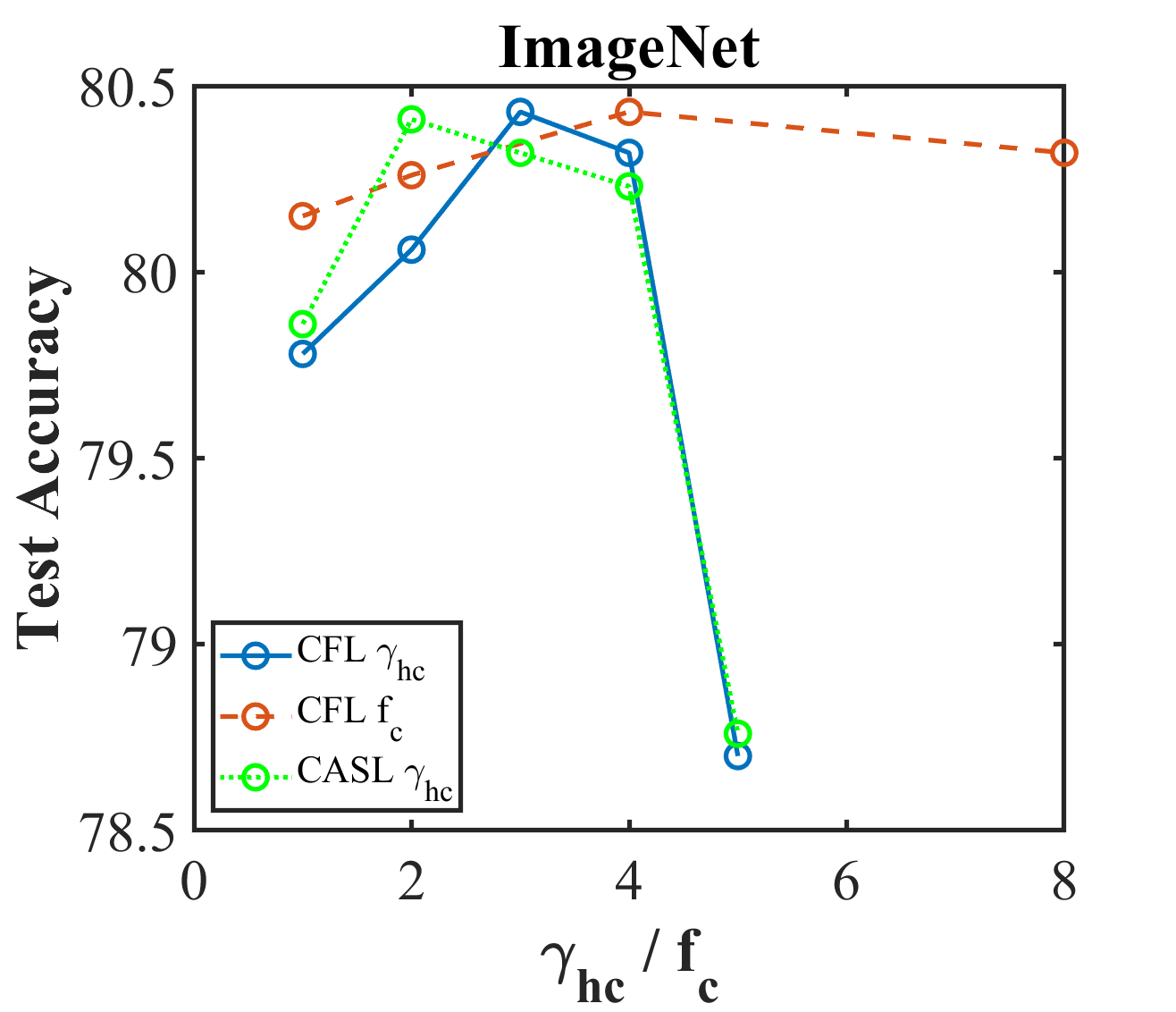}}
		\caption{\textbf{$\gamma_{hc} \ f_c$:} This figure shows the impact of $\gamma_{hc}$ and $f_c$ when using cyclical focal loss (CFL) or cyclical asymmetric focal loss (CASL) for training ImageNet with the TResNet\_m architecture.  The best values for $\gamma_{hc}$ is 2 for CASL or 3 for CFL and the accuracy rapidly drops for higher values.  However, the accuracy is relatively stable for a range of settings for the $f_c$ parameter.
		}
		\label{fig:imagenethp}
	\end{center}
	\vskip -0.2in
\end{figure}

\begin{table}[htb]
	\caption{\textbf{Imbalanced Datasets:} Comparison of the test classification accuracies for balanced and imbalanced versions with 4,000 training samples from the CIFAR-10 and CIFAR-100 datasets for training with the following loss functions: cross-entropy softmax (CE), focal loss (FL), asymmetric loss (ASL), cyclical focal loss (CFL), and cyclical asymmetric focal loss (CASL).  Here we display test dataset performance for networks trained on balanced, 5/3 imbalanced, and 6/2 imbalanced.  See the text for a description of these datasets. }
	\label{tab:imbalanced}
	\begin{center}
		\begin{tabular}{|l|c|c|c|}
			\hline
			4K &    & 5/3 & 6/2 \\
			CIFAR10 & Balanced & Imbalance & Imbalance   \\
			\hline
			CE & 86.83$\pm$ 0.24 & 86.57$\pm$ 0.04  & 85.85$\pm$ 0.07   \\
			\hline
			FL & 86.06$\pm$ 0.21& 85.96$\pm$ 0.09  & 85.30$\pm$ 0.18   \\
			\hline
			ASL & 86.23$\pm$ 0.26& 86.44$\pm$ 0.22  &  85.81$\pm$ 0.20  \\
			\hline
			CFL & \textbf{87.32$\pm$ 0.33}  & 87.03$\pm$ 0.19  &  \textbf{86.34$\pm$ 0.13}  \\
			\hline
			CASL & 87.21$\pm$ 0.57& \textbf{87.20$\pm$ 0.12} &    86.09$\pm$ 0.22 \\
			\hline
			\hline
			4K &    & 5/3 & 6/2 \\
			CIFAR100 & Balanced & Imbalance & Imbalance   \\
			\hline
			CE & 53.33$\pm$ 0.33 & 53.26$\pm$0.37  & 51.35$\pm$0.36   \\
			\hline
			FL & 50.20$\pm$ 0.46& 49.43$\pm$0.41  & 47.99$\pm$0.78   \\
			\hline
			ASL & 50.95$\pm$ 0.24& 50.91$\pm$0.31  &  49.77$\pm$0.49  \\
			\hline
			CFL & \textbf{54.13$\pm$ 0.63}  & 53.71$\pm$0.33  &  52.20$\pm$0.27  \\
			\hline
			CASL & 53.69$\pm$ 0.15& \textbf{53.74$\pm$0.40} & \textbf{52.41$\pm$0.49}    \\
			\hline
		\end{tabular}
	\end{center}
	\vskip -0.1in
\end{table}

\subsection{ImageNet}
\label{subsec:imagenet}

This Section contains the test accuracy results for training a TResNet model on ImageNet.  ImageNet is a large scale dataset, and the results here show the generality of our cyclical loss functions.
The sixth row of Table \ref{tab:losses} contains the results of training with the same five loss functions as discussed previously.
For ImageNet, each entry is the mean and the standard deviation of two runs with the same hyper-parameters and loss function. 

Once again, using the original focal loss (FL) substantially reduces the performance of the network.  In this case, using the asymmetric focal loss (ASL) regains only part of the lost performance. We expect that the loss contribution from the error or negative loss might be hurting the classification performance and a smaller $\gamma_-$ would help (i.e., $\gamma_- < 4$).

In spite of this degradation of performance, the test accuracy when training with our cyclical focal loss and the cyclical asymmetric focal loss is consistently better than the cross-entropy loss, the focal loss, or the asymmetric loss.

Figure \ref{fig:imagenet_training} shows the test accuracy curves over the course of training the network for cross-entropy softmax (CE) and our cyclical asymmetric focal loss (CASL).  The two curves look very similar but it is notable that training with CASL provides a slightly faster learning curve early in the training, which supports our intuition that cyclical focal loss helps the learning in the early epochs.  In addition, the final performance is better than CE, as is confirmed in the sixth row of Table \ref{tab:losses}.

Figure \ref{fig:imagenethp} shows the test accuracies for ImageNet over a range of values for $\gamma_{hc}$ and $ f_c$ for CFL and CASL.  For training with CFL the best value for $\gamma_{hc}$ is 3 and for CASL the best value is 2, but either the results for 2 or 3  fall within the precision of our experiments (i.e., $\pm0.1$).  
While CFL and CASL do introduce the two new hyper-parameters $\gamma_{hc}$ and $f_c$, we found that the results were insensitive to $f_c$ (by default, we used $f_c = 4$) and that $\gamma_{hc} = 2$ or 3 generally worked best, which reduces a grid search to only two tests for $\gamma_{hc}$.   

\subsection{Imbalanced Training Set}
\label{subsec:4k}

This Section evaluates test accuracy when training with cyclical focal loss and cyclical asymmetric focal loss for balanced and imbalanced versions of the CIFAR-10 and CIFAR-100 datasets when there are only 4,000 training samples.  
For the balanced version, we take the first 4,000 training samples in the CIFAR-10 dataset, which give a count per class for the ten classes (airplane, automobile, bird, cat, deer, dog, frog, horse, ship, truck) as 
396, 366, 420, 384, 422, 393, 414, 385, 407, and 413, respectively. 
While this isn't perfectly balanced due to the randomness in the order of the appearance of training samples, we consider this as our balanced dataset.  Similarly, we take the first 4,000 training samples in the CIFAR-100 dataset as our balanced dataset, although the number per class is not exactly 40 per class.

For the 5/3 imbalanced version of CIFAR-10, we provide a data sampler that specifies which samples to use so that we get a count per class for the ten classes as 
490, 470, 450, 430, 410, 390, 370, 350, 330, and 310, respectively.
Specifically, we take the first 490 examples of the first class (i.e., airplane) from the training dataset to be the samples for this class.  A similar procedure is used for the other nine classes.   The total number of training samples is 4,000, which is the same number as in the balanced version.
For CIFAR-100, it is a bit more complicated.  The 5/3 sampler starts at 50 samples per class and is reduced by $round(i / 5)$, where $i$ is the class label.  Then there is a slight adjustment at the end so there is exactly 4,000 training samples.
In the spirit of reproducibility, all our samplers are provided at \url{https://github.com/lnsmith54/CFL}.

For the 6/2 imbalanced version of CIFAR-10, we provide a data sampler that specifies which samples to use so that we get a count per class for the ten classes as 
580, 540, 500, 460, 420, 380, 340, 300, 260, and 220, respectively.
The sampling procedure for this 6 -2 imbalanced version is the same as used for the 5 -3 imbalanced version.  
For CIFAR-100, the 6/2 sampler starts at 60 samples per class and is reduced by $round(i / 2.5)$ and there is a slight adjustment at the end so there are exactly 4,000 training samples.

Table \ref{tab:imbalanced} illustrates the test accuracies comparing cross-entropy softmax (CE), focal loss (FL) \cite{lin2017focal}, asymmetric focal loss (ASL) \cite{ridnik2021asymmetric}, cyclical focal loss (CFL), and cyclical asymmetric focal loss (CASL).
As to be expected, the performance drops significantly when training with only a fraction of the CIFAR-10 or CIFAR-100 datasets.  For the balanced version,  focal loss and the asymmetric focal loss test accuracies are lower than  the accuracies from training with cross-entropy softmax.  On the other hand, training with the cyclical versions of focal loss improves on the performance over cross-entropy softmax even more clearly here than when training with the full dataset.  This provides some evidence that cyclical focal loss is particularly beneficial in few-shot or medium shot scenarios when there is only a limited amount of labeled training data.

The results for training with the 5/3 imbalanced version are in the third column of Table \ref{tab:imbalanced} and they shows a reduction in overall performance for cross-entropy softmax.  The performance for focal loss deteriorates slightly and it is still inferior to cross-entropy softmax.  However, training with the asymmetric focal loss improves on its performance compared to its performance with the balanced dataset for CIFAR-10 and stays steady for CIFAR-100.  However, the performance for our CFL and CASL is superior to the performance obtained by training with the other loss functions.

The fourth column of Table \ref{tab:imbalanced} provides the results for training with the 6/2 imbalanced version.  With the greater imbalance in the dataset, the performance of the cross-entropy loss continues to decline.  This Table shows that focal loss is still inferior to cross-entropy but the ASL approach provides comparable performance to cross-entropy for CIFAR-10 but is inferior to training with the cross-entropy loss for CIFAR-100.  However, the performance for both CFL and CASL is significantly higher than the results from training with the other loss functions.

\subsection{Open Long Tailed Recognition}
\label{subsec:OLTR}

Liu,~\etal \cite{liu2019large} defined the Open Long-Tailed Recognition (OLTR) challenge as learning from a highly imbalanced dataset, with many-shot ($\geq 100$), medium-shot ($\leq$ 100 and $>$ 20), and few-shot ($\leq$ 20) per class, plus an open-world setting in which OLTR must handle previously unseen classes.  

One expects that adding focal loss (FL) or the asymmetrical focal loss (ASL) would improve the results for the few-shot case, which was mostly confirmed in our experiments.  The question we answer here is what impact does training with the cyclical focal loss (CFL) or the cyclical asymmetric focal loss (CASL) have on these highly-imbalanced ImageNet-LT and Places-LT datasets.

Here we present the results for ImageNet-LT in Table \ref{tab:OLTR-Imagenet} and Places-LT in Table \ref{tab:OLTR-Places}  using code modified from that provided by the authors\footnote{OLTR instructions, code and data: \url{https://github.com/zhmiao/OpenLongTailRecognition-OLTR}}.  Our revised code is available at \url{https://github.com/lnsmith54/CFL}.


In Table \ref{tab:OLTR-Imagenet} and Table \ref{tab:OLTR-Places}, performance is evaluated under both the closed-set (test set contains no unknown classes) and open-set (test set contains unknown classes) settings to highlight their differences. Under each setting, in addition to the overall top-1 classification accuracy, these Tables list the accuracy of three disjoint subsets: many-shot classes (classes with over training 100 samples), medium-shot classes (classes each with between 20 and 100 training samples per class), and few-shot classes (classes with under 20 training samples).
For the open-set setting, the F-measure is also reported for a balanced treatment of precision and recall following \cite{bendale2016towards}.

\begin{table*}[htb]
	\footnotesize
	
	\centering
	\caption{\textbf{ImageNet-LT:} Comparison of the top-1 test classification accuracies on ImageNet-LT of the original algorithm to adding cyclical focal loss.  The results presented here for OLTR~\cite{liu2019large} are from the original paper.  The medium-shot and few-shot performance when using CFL or CASL exceed the results from OLTR.  }
	\vspace{5pt}
	\bgroup
	\begin{tabular}{l|cccc|cccc}
		\hline
		\textbf{Backbone Net} & \multicolumn{4}{c|}{\footnotesize{\textbf{closed-set setting}}} & \multicolumn{4}{c}{\footnotesize{\textbf{open-set setting}}} \\
		ResNet-152 & $>100$ & $\leqslant100$ \& $>20$ & $<20$ && $>100$ & $\leqslant100$ \& $>20$ & $<20$ &\\
		\textbf{Methods} & \textbf{Many-shot} & \textbf{Medium-shot} & \textbf{Few-shot} & \textbf{Average} & \textbf{Many-shot} & \textbf{Medium-shot} & \textbf{Few-shot} & \textbf{F-measure} \\
		\hline\hline
		OLTR & 43.2 & 35.1 & 18.5 & 32.3 & 41.9 & 33.9 & 17.4 & \textbf{0.474} \\
		\hline
		OLTR + FL & \textbf{46.7} & 36.5 & 16.0 & 33.1 & \textbf{45.2} & 34.8 & 15.1 & 0.446 \\
		OLTR + ASL & 45.8 & 37.6 & 18.5 & 33.9 & 44.2 & 36.3 & 17.5 & 0.450 \\
		\hline
		OLTR + CFL & 42.6 & \textbf{37.7} & \textbf{23.3} & \textbf{34.5} & 40.0 & \textbf{35.4} & \textbf{21.7} & 0.445 \\
		OLTR + CASL & 42.5 & 37.4 & 22.8 & 34.2 & 39.8 & 35.1 & 21.2 & 0.441 \\
		\hline
	\end{tabular}\vspace{-5pt}
	\egroup
	\label{tab:OLTR-Imagenet}
\end{table*}

\begin{table*}[htb]
	\footnotesize
	
	\centering
	\caption{\textbf{Places-LT:}Comparison of the top-1 test classification accuracies on Places-LT of the original algorithm to adding cyclical focal loss.  The results presented here for OLTR~\cite{liu2019large} are from the original paper.   The medium-shot, few-shot, and average performance when using CFL or CASL exceed the results from OLTR.}
	\vspace{5pt}
	\bgroup
	\begin{tabular}{l|cccc|cccc}
		\hline
		\textbf{Backbone Net} & \multicolumn{4}{c|}{\footnotesize{\textbf{closed-set setting}}} & \multicolumn{4}{c}{\footnotesize{\textbf{open-set setting}}} \\
		ResNet-152 & $>100$ & $\leqslant100$ \& $>20$ & $<20$ && $>100$ & $\leqslant100$ \& $>20$ & $<20$ &\\
		\textbf{Methods} & \textbf{Many-shot} & \textbf{Medium-shot} & \textbf{Few-shot} & \textbf{Average} & \textbf{Many-shot} & \textbf{Medium-shot} & \textbf{Few-shot} & \textbf{F-measure} \\
		\hline\hline
		OLTR & \textbf{44.7} & 37 & 25.3 & 35.9 & \textbf{44.6} & 36.8 & 25.2 & 0.464 \\
		\hline
		OLTR + FL & 43.4 & 39.7 & 27.1 & 36.7 & 43.3 & 39.4 & 26.8 & 0.493 \\
		OLTR + ASL & 44.4 & 39.9 & 28.4 & 37.6 & 44.2 & 39.5 & 27.9 & 0.499 \\
		\hline
		OLTR + CFL & 43.0 & \textbf{40.7} & 30.8 & \textbf{38.2} & 42.5 & \textbf{40.2} & 30.0 & \textbf{0.502} \\
		OLTR + CASL & 43.7 & 39.8 & \textbf{31.9} & \textbf{38.5} & 43.4 & 39.3 & \textbf{31.3} & \textbf{0.503} \\
		\hline
	\end{tabular}\vspace{-5pt}
	\egroup
	\label{tab:OLTR-Places}
\end{table*}


The second rows of Table \ref{tab:OLTR-Imagenet} (for ImageNet-LT) and Table \ref{tab:OLTR-Places} (for Places-LT) show the results for the OLTR from the paper \cite{liu2019large}.
The third and fourth rows show that including either focal loss or the asymmetrical focal loss often improves the results for the medium-shot and few-shot cases, with a small degradation of the many-shot performance.

The fifth and sixth rows of both Tables show the performance results of using our CFL or CASL loss functions in place of cross-entropy softmax in the OLTR methodology.
It is noteworthy that both cyclical focal loss and the cyclical asymmetric focal loss improve the results for the medium-shot and few-shot cases relative to the focal loss or the asymmetrical focal loss 
(with a small degradation of the many-shot performance). The best overall accuracy in the closed-set setting  for both datasets was obtained by using the CFL and CASL losses.  In the open-set setting, the best accuracies for the medium-shot and low-shot in both datasets were obtained when training with our CFL and CASL loss function and for Places-LT, these loss functions obtained the highest F-measure.

These experiments demonstrate the superiority of cyclical focal loss and cyclical asymmetric focal loss to cross-entropy softmax and focal loss when the input data is highly imbalanced.

\section{Conclusions}
\label{sec:conclusions}

In this work, we introduce two novel loss functions: the cyclical focal loss (CFL) and the cyclical asymmetric focal loss (CASL). These loss functions add a new loss term to the focal loss that more heavily weights confident training samples in the first epochs of a neural network's training.  As the training progresses, the focal loss term that weights less confident samples dominates the loss.   In the final epochs, the loss function returns to fine-tuning on the confident samples learned over the course of the training.

Our extensive empirical analysis demonstrates that CFL and CASL provide comparable or superior performance to cross-entropy softmax, focal loss, and asymmetric focal loss across balanced, imbalanced, and long-tailed datasets. We did not find CFL or CASL harmful to the performance relative to training with cross-entropy softmax or focal loss.  In addition, CFL and CASL demonstrated superior performance in the case of limited labeled training data where our experiments showed the results from training with only 4,000 training samples from the CIFAR-10 and CIFAR-100 datasets.  Furthermore, our experiments show that our results are robust to the choice of new hyper-parameters: a default choice of $f_c = 4$ and $\gamma_{hc} = 2$ or 3 worked well across all of our experiments.

Our experiments provide evidence that our cyclical focal loss and cyclical asymmetric focal loss are more universal loss functions over more scenarios and applications than cross-entropy softmax or the focal loss functions. 
Therefore, they can be used as drop in replacements for cross-entropy softmax and are especially beneficial when there are a limited number of labeled training samples or there is imbalance in the number of samples in each class.

\section*{Acknowledgements}
We thank the US Office of Naval Research for their support of this research.  
The views, opinions and/or findings expressed are those of the authors and do not reflect the official policy or position of the US Navy, Department of Defense or the US Government.

\bibliographystyle{unsrt}  
\bibliography{CFL}

\newpage
\appendix
\onecolumn

\section{Software and Implementation}

We modified the implementation of the asymmetric focal loss \cite{ridnik2021asymmetric} to include a cyclical focal loss option.  The original code is available at \url{https://github.com/Alibaba-MIIL/ASL}.  The asymmetric focal loss provided the flexibility to compare cyclical versions to both the original and the asymmetric focal losses.  Specifically, we modified the file from the original asymmetric focal loss at  \url{src/loss\_functions/losses.py} and made a copy of the class \texttt{class~ASLSingleLabel} that we named \texttt{class~Cyclical\_FocalLoss}.  We then added the following code to convert this into cyclical focal loss:
\begin{lstlisting}
	if  self.factor*epoch < self.epochs:
	    eta = 1.0 -  self.factor *epoch/(self.epochs-1)
	elif self.factor == 1:
	    eta = 0.0
	else:
	   eta =  (self.factor*epoch/(self.epochs-1) - 1.0)/(self.factor - 1.0)
	   
	positive_w = torch.pow(1 + xs_pos,self.gamma_hc * targets)
	log_preds = log_preds * ((1 - eta)* asymmetric_w + eta * positive_w)
\end{lstlisting}
The full revised code is available at \url{https://github.com/lnsmith54/CFL}, such as in \texttt{timm/loss/asl\_focal\_loss.py}.

In addition, we used PyTorch Image Models (timm) \cite{rw2019timm} as a framework in our experiments on CIFAR and ImageNet.  This framework provides the models and downloads the data used in our our experiments.  
The original code is available at \url{https://github.com/rwightman/pytorch-image-models}.  The file \texttt{train.py} was modified by inserting several new input parameters via calls to \texttt{add\_argument}, modifying the loss setup to call our focal loss code for FL, ASL, CFL, or CASL. 
There were additional modifications made to add a sampler that used only part of a dataset, which was used in our experiments with limited labeled training data. 

In summary, there were a number of modifications made in order to read in hyper-parameters, to call the cyclical focal loss, and to expedite running our experiments.  The full revised \texttt{train.py} is available as part of code base at \url{https://github.com/lnsmith54/CFL}.

Also, we made use of the code provided for the open long tailed recognition experiments \cite{liu2019large}.
The original code is available at \url{https://github.com/zhmiao/OpenLongTailRecognition-OLTR}.  We copied  the same file as described above for CFL to this \texttt{loss/} folder.  We also created configuration files that used FL, ASL, CFL, or CASL instead of the default softmax loss.  Finally, we modified \url{main.py} to set $\gamma_{hc}$ for a given experiment more easily.   Again,  there were a number of modifications made in order to read in hyper-parameters, to call the cyclical focal loss, and to expedite our experiments. 
The full revised code is available at \url{https://github.com/lnsmith54/CFL}.

\begin{table*}[tb]
	\caption{The hyper-parameters used for experiments whose results are presented in the main paper.}
	\label{tab:HP}
	\begin{center}
		\begin{small}
			\begin{sc}
				\begin{tabular}{|l|c|c|c|c|c|c|c|c|}
					\hline
					Table & Dataset & Model & Batch size & LR & WD & CFL $\gamma_{hc}$ & CASL $\gamma_{hc}$ & $f_c$ \\
					\hline
					Table \ref{tab:losses} & CIFAR-10 & TResNet\_m  & 384 & 0.15 & $5 \times 10^{-4} $ & 3 & 2 & 4 \\
					\hline
					Table \ref{tab:losses} & CIFAR-10 & ResNet50 & 192 & 0.3 & $5 \times 10^{-4} $ & 3 & 2 & 4 \\
\hline
					Table \ref{tab:losses} & CIFAR-10 & Efficient\_B0 & 192 & 0.2 & $2 \times 10^{-4} $ & 4 & 3 & 4 \\
\hline
					Table \ref{tab:losses} & CIFAR-100 & TResNet\_m  & 64 & 0.1 & $2 \times 10^{-4} $ & 3 & 3 & 4 \\
\hline
					Table \ref{tab:losses} & ImageNet & TResNet\_m  & 192 & 0.6 & $2 \times 10^{-5} $ & 3 & 2 & 4 \\
\hline
					Table \ref{tab:imbalanced} & CIFAR-10 & Balanced  & 64 & 0.3 & $5 \times 10^{-5} $ & 3 & 3 & 4 \\
\hline
					Table \ref{tab:imbalanced} & CIFAR-10 & 5-3 Sampler  & 64 & 0.4 & $2 \times 10^{-5} $ & 3 & 2 & 4 \\
\hline
					Table \ref{tab:imbalanced} & CIFAR-10 & 6-2 Sampler  & 64 & 0.4 & $2 \times 10^{-5} $ & 2 & 2 & 4 \\
\hline
					Table \ref{tab:imbalanced} & CIFAR-100 & Balanced  & 64 & 0.3 & $5 \times 10^{-5} $ & 3 & 3 & 4 \\
					\hline
					Table \ref{tab:imbalanced} & CIFAR-100 & 5-3 Sampler  & 64 & 0.4 & $2 \times 10^{-5} $ & 4 & 4 & 4 \\
					\hline
					Table \ref{tab:imbalanced} & CIFAR-100 & 6-2 Sampler  & 64 & 0.4 & $2 \times 10^{-5} $ & 2 & 4 & 4 \\
					\hline
				\end{tabular}
			\end{sc}
		\end{small}
	\end{center}
	\vskip -0.1in
\end{table*}

\section{Command Lines and Hyper-parameters}

In the spirit of easy replication, it is important to know the values of the hyper-parameters used.  Table \ref{tab:HP} specifies the batch sizes, learning rates, and weight decay values used for the results in Table \ref{tab:losses} and Table \ref{tab:imbalanced} in the main body of the paper.

The PyTorch Image Models implementation provides for command line input of a large number of hyper-parameters.  Here we specified the command line used in the experiments, and all other hyper-parameters used the default settings provided by the framework.  

For CIFAR-10 the following command line is an example of what was used for the results in Table \ref{tab:losses}:
\begin{verbatim}
	CUDA_VISIBLE_DEVICES=0 python train.py  data/cifar10 --dataset torch/cifar10
	-b 384 --model tresnet_m --checkpoint-hist 4 --sched cosine --epochs 200 --lr 0.15 
	--warmup-lr 0.01 --warmup-epochs 3 --cooldown-epochs 1 --weight-decay 5e-4 
	--amp --remode pixel --reprob 0.6 --aug-splits 3 --aa rand-m9-mstd0.5-inc1 
	--resplit --split-bn  --dist-bn reduce --focal_loss asym-cyclical --gamma_hc 3 
	--gamma_pos 2 --gamma_neg 2 --cyclical_factor 4
\end{verbatim}
Table \ref{tab:HP} reflects any changes to the hyper-parameters that would be made in this command line in various experiments.  In addition, we used a cosine learning rate schedule with a warmup during training, as indicated by the \url{--sched~ cosine} parameter.

Default values for hyper-parameters not specified in this command line were used (i.e., see the software at \url{https://github.com/lnsmith54/CFL} or the original code at \url{https://github.com/rwightman/pytorch-image-models} for default values of the hyper-parameters).  The new hyper-parameters are: focal\_loss, gamma\_hc, gamma\_pos,  gamma\_neg, and cyclical\_factor.  The focal\_loss parameter specifies whether to use the loss: if set to ``asym-cyclical'' it calls the cyclical asymmetric focal loss function; if set to ``asym'' it calls the asymmetric focal loss; and if set to anything else, then cross-entropy softmax is called.  The other parameters (gamma\_hc, gamma\_pos,  gamma\_neg, and cyclical\_factor) correspond to $\gamma_{hc}, \gamma_+, and \gamma_-$, respectively.

For CIFAR-100, the following command line was the same as the one for CIFAR-10 except to replace cifar10 with cifar100 as follows:
\begin{verbatim}
	CUDA_VISIBLE_DEVICES=0 python train.py  data/cifar100 --dataset torch/cifar100 
\end{verbatim}
Table \ref{tab:HP} reflects any changes to the hyper-parameters. 

An example command line for submitting an experiment on with Imagenet to produce the results in Table \ref{tab:losses}:
\begin{verbatim}
	./distributed_train.sh 4 data/imagenet -b=192 --lr=0.6 --warmup-lr 0.02 
	--warmup-epochs 3 --weight-decay 2e-5 --cooldown-epochs 1 
	--model-ema --checkpoint-hist 4 --workers 8 --aa=rand-m9-mstd0.5-inc1 -j=16 --amp 
	--model=tresnet_m --epochs=200 --mixup=0.2 --sched='cosine' --reprob=0.4 
	--remode=pixel --focal_loss asym-cyclical --gamma_hc 3 --gamma_pos 2 --gamma_neg 2 
	--cyclical_factor 4
\end{verbatim}

The experiments for the Large-Scale Long-Tailed Recognition in an Open World (OLTR) challenge followed the instructions given at \url{https://github.com/zhmiao/OpenLongTailRecognition-OLTR}.  The software is not set up to read the hyper-parameters so the program defaults were used and no fine tuning was attempted.  Specifically, the command lines for running OLTR are:
\begin{verbatim}
#  ImageNet-LT
# Stage 1 training:
CUDA_VISIBLE_DEVICES=0,1,2,3 python main.py --config ./config/ImageNet_LT/stage_1.py 

# Stage 2 training:
CUDA_VISIBLE_DEVICES=0 python main.py 
	--config ./config/ImageNet_LT/stage_2_meta_embedding.py

# Close-set testing:
CUDA_VISIBLE_DEVICES=1 python main.py 
	--config ./config/ImageNet_LT/stage_2_meta_embedding.py --test

# Open-set testing (thresholding)
CUDA_VISIBLE_DEVICES=2 python main.py 
	--config ./config/ImageNet_LT/stage_2_meta_embedding.py --test_open

# Test on stage 1 model
CUDA_VISIBLE_DEVICES=3 python main.py 
	--config ./config/ImageNet_LT/stage_1.py --test 
\end{verbatim}
We modified the file \texttt{main.py} to permit reading in values for our new hyper-parameters focal\_loss, gamma\_hc, gamma\_pos,  gamma\_neg, and cyclical\_factor.  We modified \texttt{run\_networks.py} to call focal loss, asymmetric focal loss, or our cyclical focal loss versions.  The modified code is available at \url{https://github.com/lnsmith54/CFL}.



\end{document}